\pdfoutput=1

\documentclass[11pt]{article}

\usepackage[preprint]{coling}
\usepackage{times}
\usepackage{latexsym}

\usepackage[T1]{fontenc}

\usepackage[utf8]{inputenc}

\usepackage{microtype}

\usepackage{inconsolata}

\usepackage{graphicx}
\usepackage{subcaption}
\usepackage{enumitem}
\usepackage{booktabs}
\usepackage{multirow}
\usepackage{makecell}
\usepackage{amsmath}
\usepackage{bbding}
\usepackage{algorithm}
\usepackage{algpseudocode}
\usepackage{xcolor}
\usepackage{microtype}
\usepackage{colortbl}
\usepackage{pifont}
\usepackage{url}
\usepackage{adjustbox}
\usepackage[listings,skins,breakable]{tcolorbox}
\usepackage{multicol}
\definecolor{mycell}{gray}{.95}
\definecolor{mycelltwo}{RGB}{255, 182, 193}
\definecolor{ForestGreen}{rgb}{0.13, 0.55, 0.13}
\definecolor{DarkCoral}{rgb}{0.72, 0.33, 0.31}
\definecolor{lightgray}{RGB}{230, 230, 230}
\newcommand{\ie}{\textit{i.e.}}
\newcommand{\eg}{\textit{e.g.}}
\newcommand{\etc}{\textit{etc.}}

\newcommand{\symboldraft}{\raisebox{-0.2pt}{\includegraphics[scale=0.7]{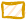}}}
\newcommand{\symboltarget}{\raisebox{-0.2pt}{\includegraphics[scale=0.6]{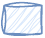}}}
\newcommand{\symbollatency}{\raisebox{-0.2pt}{\includegraphics[scale=1.0]{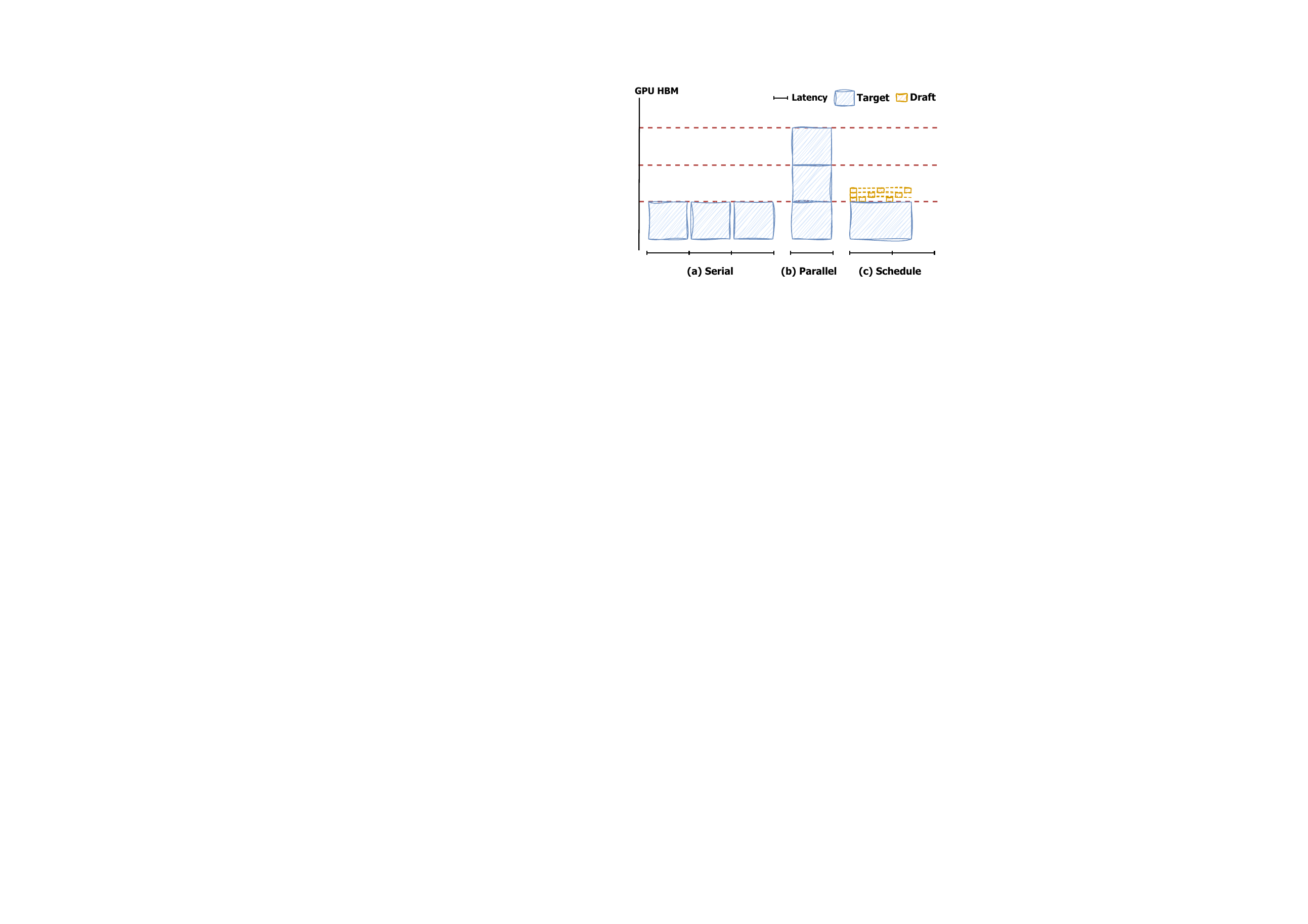}}}
\newcommand{\symboldraftone}{\raisebox{-0.2pt}{\includegraphics[scale=0.7]{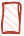}}~}
\newcommand{\symboldrafttwo}{\raisebox{-0.2pt}{\includegraphics[scale=0.7]{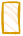}}~}
\newcommand{\symboldraftthree}{\raisebox{-0.2pt}{\includegraphics[scale=0.7]{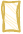}}~}
\newcommand{\symbolverify}{\raisebox{-0.2pt}{\includegraphics[scale=0.7]{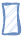}}~}
\newcommand{\symbolresample}{\raisebox{-0.2pt}{\includegraphics[scale=0.7]{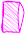}}~}
\newcommand{\symboldraftmodelone}{\raisebox{-0.2pt}{\includegraphics[scale=0.2]{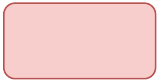}}~}
\newcommand{\symboldraftmodeltwo}{\raisebox{-0.2pt}{\includegraphics[scale=0.2]{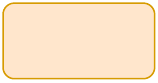}}~}
\newcommand{\symboldraftmodelthree}{\raisebox{-0.2pt}{\includegraphics[scale=0.2]{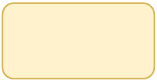}}~}
\newcommand{\symboltargetmodel}{\raisebox{-0.2pt}{\includegraphics[scale=0.2]{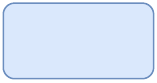}}~}
\newcommand{\yes}{\color{green!60!black}\ding{51}}
\newcommand{\no}{\color{red!60!black}\ding{55}}
\newcommand{\triangleSymbol}{\textcolor{blue!60!black}{\ding{115}}}
\newcommand{\squareSymbol}{\textcolor{orange!60!black}{\ding{110}}}
\newtcolorbox{titleEnv}{
colframe=black!80,
colback=gray!10,
fonttitle=\bfseries,
coltitle=black,
left=3pt,
right=3pt,
top=3pt,
bottom=3pt,
boxrule=0.4mm,
arc=3mm
}

\title{
\textsc{SeeD}: Accelerating Reasoning Tree Construction via Scheduled Speculative Decoding
}

\author{Zhenglin Wang\thanks{~~Equal Contribution.},\hspace{1.5mm}
   Jialong Wu$^{*}$,\hspace{1.5mm}
   Yilong Lai,\hspace{1.5mm}
   Congzhi Zhang,\hspace{1.5mm}
   Deyu Zhou\thanks{~~Corresponding Author.}\hspace{1.5mm}
   \\
        \hspace{0.5mm}School of Computer Science and Engineering, Key Laboratory of Computer Network\\
        and Information Integration, Ministry of Education, Southeast University, China
    \\
        \texttt{\{zhenglin, jialongwu, yilong.lai, zhangcongzhi, d.zhou\}@seu.edu.cn} \\
}

\begin{document}
\maketitle
\begin{abstract}
Large Language Models (LLMs) demonstrate remarkable emergent abilities across various tasks, yet fall short of complex reasoning and planning tasks.
The tree-search-based reasoning methods address this by encouraging the exploration of intermediate steps, surpassing the capabilities of chain-of-thought prompting.
However, significant inference latency is introduced due to the systematic exploration and evaluation of multiple thought paths.
This paper introduces \textsc{SeeD}, a novel and efficient inference framework to improve both runtime speed and GPU memory management concurrently.
Based on a scheduled speculative execution, \textsc{SeeD} efficiently handles multiple iterations for thought generation and state evaluation, leveraging a rounds-scheduled strategy to manage draft model dispatching.
Extensive experimental evaluations on three reasoning datasets demonstrate the superior speedup performance of \textsc{SeeD}\footnote{\raisebox{-2.2pt}{\includegraphics[scale=0.03]{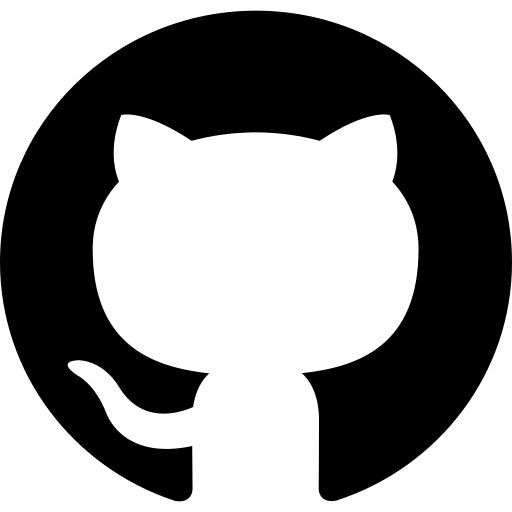}}~~\url{https://github.com/Linking-ai/SEED}}.
\end{abstract}

\section{Introduction}
Despite Large Language Models (LLMs) have shown remarkable emergent abilities across a variety of tasks~\citep{ouyang2022training,chatgpt,touvron2023llama, touvron2023llama2,achiam2023gpt}, their performance on the complex reasoning and planning tasks remains suboptimal~\citep{zhang-etal-2024-star}.
Traditional or simple prompting techniques~\citep{wei2022chain, kojima2022large}, which have been widely leveraged, are insufficient for the tasks that require exploratory actions or strategic lookahead~\cite{liao2024look}.

\begin{figure}[t]
    \centering
    \includegraphics[width=0.45\textwidth]{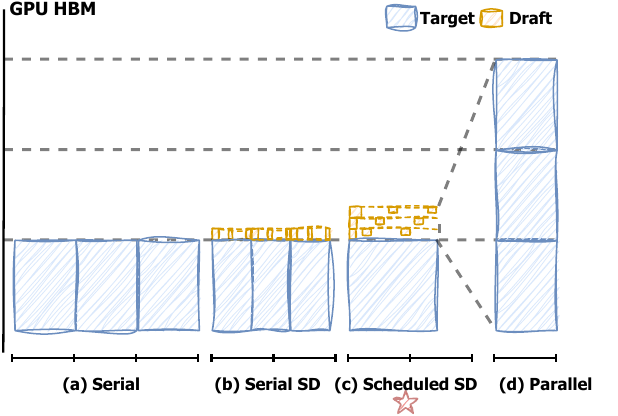}
    \caption{Illustration of four LLM execution strategies for generating 3 sequences in Reasoning Tree construction:
    (a) \textit{Serial}, where executions are operated  one after another, simplifying resource management but increasing overall execution time;
    (b) \textit{Seiral SD}, where speculative decoding is used for each execution; 
    (c) \textit{Scheduled SD}, which involves several parallel draft models and one target model;
    (d) \textit{Parallel}, where multiple executions run concurrently, reducing completion time but increasing GPU HBM.
    \symboltarget ~refers to a large target model,
    \symboldraft ~signifies a smaller draft model,
    \symbollatency ~represents \texttt{a unit length} of execution time.
    }
    \label{fig:intro}
\end{figure}

Tree-Search-Based (TSB) reasoning methods effectively harness the planning and reasoning capabilities of LLMs by decomposing the problems and subsequently orchestrating a structured plan~\citep{hui2024rot}.
These methods not only leverage the inherent strengths of LLMs in processing vast datasets but also address their limitations in dynamic problem-solving scenarios~\citep{hao2023reasoning, guan2023leveraging}.
For example, \citet{yao2024tree} introduced Tree-of-Thoughts (ToT) prompting, which generalizes beyond Chain-of-Thought (CoT) prompting by fostering the exploration of intermediate thoughts that serve as crucial steps in general problem-solving with LLMs.
Following this way, subsequent works, such as Reasoning via Planning (RAP)~\citep{hao2023reasoning} and Refection on search Trees (RoT) are proposed~\citep{hui2024rot}.
These approaches leverage the capabilities of LLMs to generate and evaluate the intermediate thoughts and then integrate them with search algorithms to improve the problem-solving efficiency.

However, such methods introduce a serious issue of inference latency due to the requirement for systematic exploration of thoughts with lookahead and backtracking.
TSB reasoning methods primarily consist of two key parts, tree construction and the search algorithm.
Recent studies have enhanced the efficiency of the search algorithms by incorporating diversity rewards or pruning techniques~\citep{yan2024mirror, hui2024rot}.
To the best of our knowledge, no prior work explored the acceleration of tree construction, which is the focus of this paper.

Traditional \textit{Sequential} execution of LLMs necessitates repeated executions, leading to long execution time, as shown in Figure~\ref{fig:intro} (a).
For instance, when applying ToT prompting to execute a single sample in the GSM8K dataset, the average total runtime is approximately 100 seconds using \textit{sequential} processing with a 7B model on consumer GPUs.
If the execution of LLMs shifts from \textit{sequential} to \textit{parallel} processing, it could pose challenges for end-users or researchers only with consumer GPUs, as illustrated in Figure~\ref{fig:intro} (d).
Such condition typically exacerbates the issues related to hardware limitations, necessitating strategies for efficient resource management and optimization. 
Speculative decoding is now widely used to accelerate inference~\citep{xia-etal-2024-unlocking}, which involves employing a small draft model with a larger target model, as depicted in Figure~\ref{fig:intro} (b).
Intuitively, these draft models achieve rapid inference speeds owing to their small size.
If they are executed in parallel, concerns about the GPU memory constraints become negligible, allowing for the speed performance comparable to the scenarios illustrated in Figure~\ref{fig:intro} (d).
Moreover, speculative decoding employs a \textit{draft-then-verify} two-stage paradigm, and the target model is not fully utilized when the acceptance rate of drafted tokens is relatively high.
By increasing the number of draft models, the potential of a single target model can be effectively harnessed, ensuring its capacity is optimally utilized.

Therefore, we propose a novel and efficient inference framework, \textsc{SeeD}, to address both runtime speed and GPU memory resource management concurrently in reasoning tree construction.
\textsc{SeeD} effectively handles two scenarios: 
(1) executing multiple iterations with the same prompt; 
(2) evaluating multiple iterations with different prompts.
We utilize scheduled speculative decoding to manage the scheduling of parallel draft models.
As depicted in Figure~\ref{fig:intro} (c), given that there is only one shared target model, which can not simultaneously verify multiple draft models, we address this limitation by drawing inspiration from process scheduling in operating system management~\citep{zhao1989performance, siahaan2016comparison}.
To this end, the Rounds-Scheduled strategy which uses a First-Come-First-Serve (FCFS) queue, is employed to control and maintain the overall execution flow.

\textsc{SeeD} achieves excellent speed performance on three reasoning and planning datasets: GSM8K,  Creative Writing and Blocksworld.
It also provides a viable path for conducting \textit{batched inference} in training-free speculative decoding while preserving the original distribution, ensuring a \textbf{lossless} outcome.

Our contribution can be summarized as follows:
\begin{itemize}[itemsep=0.5pt, parsep=0pt]
\vspace{-2mm}
    \item An efficient inference framework, \textsc{SeeD}, is proposed to accelerate the both Thought Generator and State Evaluator in reasoning tree construction.
    \item Speculative Scheduled Execution that integrates parallel drafting with speculative decoding is proposed, employing an effective Rounds-Scheduled strategy to manage parallel drafting devoid of verification conflicts.
    \item Empirically, extensive experiments and analysis studies are conducted to demonstrate the effectiveness of \textsc{SeeD}. \textsc{SeeD} achieves 1.1$-$1.5$\times$ speedups, generating up to \textbf{20 additional tokens per second} across three reasoning datasets.
\end{itemize}

\section{Related Work}
\subsection{Tree-Search-Based Reasoning}
Recently, TSB reasoning methods have been widely leveraged to augment the reasoning capabilities of LLMs such as RAP~\citep{hao2023reasoning}, ToT~\citep{yao2024tree}, RoT~\citep{hui2024rot}.
These methods craft a reasoning tree allowing consider multiple reasoning paths and self-evaluate the choices to determine the next course of action.
At each reasoning step, the popular tree search algorithms such as Breadth-First Search (BFS)~\citep{bundy1984breadth} and Monte-Carlo Tree Search (MCTS)~\citep{kocsis2006bandit} are integrated to explore the tree in search of an optimal state.
Also, the construction or search of the tree requires more iterations than single sampling methods (\eg, Input-output prompting and CoT~\citep{wei2022chain}), leading to higher inference latency.
To address this, some studies introduce diversity rewards~\citep{yan2024mirror} or pruning techniques~\citep{hui2024rot} to mitigate inefficient searches during iterations, improving search efficiency.
However, these methods still overlook the inference latency caused by the iterative process of tree construction.
Instead, we focus on tree construction, leveraging speculative scheduled decoding to accelerate the process and reduce inference latency.

\subsection{Parallel Decoding}
The inference latency of LLMs has emerged as a substantial obstacle, restricting their remarkable reasoning capabilities in downstream tasks~\citep{xia-etal-2024-unlocking}.
One major factor contributing to the high inference latency is the sequential decoding strategy for token generation adopted by almost all LLMs~\citep{lu2024padellm}.
There are numerous studies have explored this challenge through parallel decoding strategies, such as Speculative Decoding (SD)~\citep{zhou2023distillspec, cai2024medusa}, Early Exiting (EE)~\citep{del2023skipdecode, elhoushi-etal-2024-layerskip}, and Non-AutoRegressive (NAR)~\citep{ghazvininejad2019mask, lu2024encode}.
In this paper, we focus on the study of Speculative Decoding.
Within SD, one line of work falls into the training-free category~\citep{sun2024spectr, liu2023online}.
This plug-and-play approach seamlessly integrates with other modular inference methods (\eg, CoT, TSB), significantly enabling direct inference acceleration and reducing inference latency on open-source models.
As far as we know, we are the \textbf{first} to explore a scheduled SD execution to integrate with the TSB framework, without modifying LLM architecture or requiring additional training and maintaining lossless output.

\section{Preliminaries}
\subsection{Speculative Decoding}
\label{sec:pre:sd}
The core technique of speculative decoding involves using a small draft model to generate tokens sequentially, with a larger target model validating these tokens~\citep{leviathan2023fast}.
Specifically, let $c$ be the input tokens, $M_{d}$ and
$M_{t}$ be the draft and the target model respectively, and $k$ be the number of draft tokens generated per step.
Speculative decoding is a \textit{Draft-then-Verify} two-stage decoding paradigm.
\footnote{In the following paper, we define ``Verification'' as the ``\textit{Verify}'' mentioned here, which includes both the verify and resampling phases.}
In the draft stage, $M_{d}$ samples a draft sequence of tokens autoregressively, denoted as $\hat{x}_1,\ldots,\hat{x}_{k}$, where $\hat{x}_i \sim p_{d}(x|\hat{x}_1,\ldots,\hat{x}_{i-1}, c)$ for $i = 1, \ldots, k$.
In the verification stage, the draft sequence of tokens along with $c$, are passed to $M_{t}$ to obtain their output distribution $p_{t}(x|\hat{x}_1,\ldots,\hat{x}_{i-1}, c)$ in parallel, and then verified from $\hat{x}_1$ to $\hat{x}_{k}$.
The draft token $\hat{x}_i$ is accepted with the probability $\min(1, \frac{p_{t}(x|\hat{x}_1,\ldots,\hat{x}_{i-1}, c)}{p_{d}(x|\hat{x}_1,\ldots,\hat{x}_{i-1}, c)})$.
Once a token is rejected, the verifying terminates and a resampling phase follows to return a new token by $M_{t}$. 
This new token is then used as the end-generated point following the accepted tokens.
As is proven in~\citet{leviathan2023fast}, this method is equivalent to sampling directly from the target LLM.
\textsc{SeeD} adopts this method, ensuring that the distribution of the generated text \textbf{remains unchanged} for both the greedy and non-greedy settings.

\subsection{Tree Attention}
\label{sec:preliminaries}
Current speculative decoding studies have demonstrated that when the draft model samples multiple candidates per position in the draft sequence, the expected acceptance length per step can be enhanced during the verification stage~\citep{chen2023accelerating}.
Additionally, the tree attention technique enables multiple candidate draft sequences to share the caches of generated tokens, further improving the efficiency of the verification stage~\citep{cai2024medusa}.
By utilizing tree attention, the verification acceptance of speculative decoding is increased.
We illustrate the detailed tree attention mask strategy in Appendix~\ref{app:tree_attention}.
Our proposed \textsc{SeeD} can leverage this approach to achieve further speedup.

\subsection{TSB Task Formulation}
\label{sec:taskformulation}
\begin{figure}[t]
    \centering
    \includegraphics[width=0.45\textwidth]{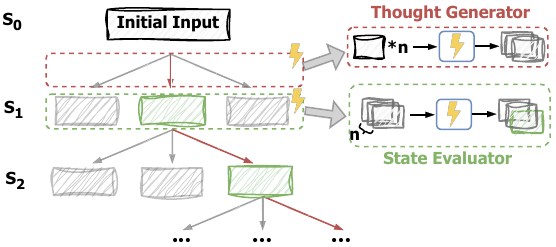}
    \caption{Two main components in reasoning tree construction, which are Thought Generator and State Evaluator, respectively.
    }
    \label{fig:framework}
\end{figure}

Given an initial input question $\mathcal{I}$, a reasoning tree is constructed with the relatively common search algorithm BFS following~\citet{yao2024tree}, as shown in Figure~\ref{fig:framework}.
In the constructed reasoning tree, each node represents a distinct state $S_i$, which includes a partial solution with the input $c$ and the progressively elaborated thoughts proposal $z_{1}, \cdots, z_{n}$.
During the expansion of each node, the Thought Generator $\textsc{G}(\cdot)$ produces multiple reasoning paths to decompose the intermediate process from the current state.
Once these thoughts are generated, the State Evaluator $\textsc{E}(\cdot)$ assesses the contribution of each path toward solving the problem, serving as a heuristic for guiding the search algorithm.
This evaluation aids in determining which states to continue exploring and in establishing the order of exploration.
Taking the root node $S_0$ as an example in Figure~\ref{fig:framework}, it first generates $n$ reasoning paths based on the same input $c$, which is the initial prompt $\mathcal{I}$ and subsequently selects the middle path by the State Evaluator for these $n$ paths.

\section{Method}

Our proposed \textsc{SeeD} is an efficient inference framework designed to accelerate the construction of a reasoning tree.
Different generation executions in the Thought Generator or the State Evaluator are conducted in distinct branches, ensuring that they do not interfere with each other.
Consequently, the Speculative Scheduled Execution is implemented in both the Thought Generator and the State Evaluator, enabling parallel processing to accelerate the overall reasoning tree construction, as detailed in Algorithm~\ref{alg:seed_bfs}.

We first introduce two phases in the Speculative Scheduled Execution in \S\ref{sec:speculative _scheduled_execution}.
Subsequently, we depict the Rounds-Scheduled Strategy designed to effectively manage parallel drafting without conflicts in \S\ref{sec:rounds_scheduled}.
The technical principle of \textsc{SeeD} is inspired by the operation system schedule.
The detailed analogy between the operation system scheduling with \textsc{SeeD} is presented in \S\ref{app:analogy}.
Finally, the combined algorithm is elaborated in \S\ref{sec:algorithm}.

\subsection{Speculative Scheduled Execution}
\label{sec:speculative _scheduled_execution}

\begin{figure*}[t]
    \centering
    \includegraphics[width=0.85\textwidth]{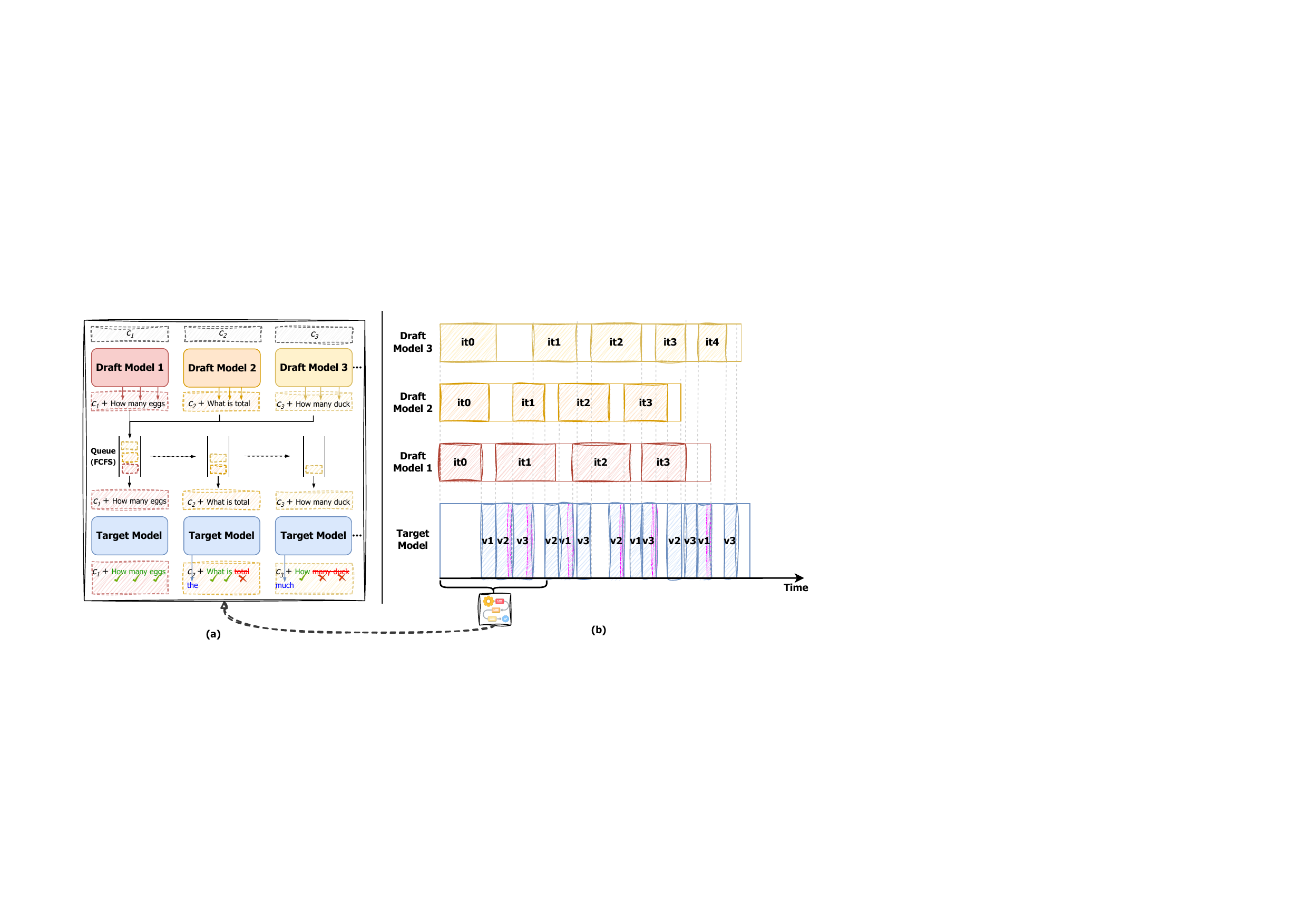}
    \caption{
    (a) The scenario where the target model manages the verification of target models at the beginning;
    (b) Overall scheduling diagram for one target model and three draft models.
    \symboldraftmodelone, \symboldraftmodeltwo, \symboldraftmodelthree represent Draft Model 1, Draft Model 2, Draft Model 3, respectively.
    \symboldraftone, \symboldrafttwo, \symboldraftthree denotes the execution times of drafting for each corresponding draft model.
    \symboltargetmodel refers to Target Model.
    \symbolverify represents the execution time of the verification phase, while \symbolresample specifies the resampling time in cases of rejection.
    }
    \label{fig:method}
    \vspace{-5mm}
\end{figure*}

We further detail the speculative scheduled execution algorithm within \textsc{SeeD}.
To enhance clarity, we delve the algorithm into two phases: the \textit{parallel drafting phase} and the \textit{sequential verification phase}.

\noindent \paragraph{Parallel Drafting Phase}
The model size significantly impacts memory usage and inference time.
In light of this, given the small size and rapid inference speed of the draft models, we can directly initialize multiple draft models corresponding to the number of thoughts, enabling parallel processes.
To be specific, if the number of thoughts $N_t$ is set to $n$, the draft models $M_{d_{1}}, M_{d_{2}}, \cdots, M_{d_{n}}$ take $c_1, c_2, \cdots, c_n$ as input tokens respectively in the drafting phase.
Note that, during the Thought Generation, the input instructions are the same, \ie, $c_1 = c_2 = \cdots = c_n$; 
during the State Evaluation, they may differ, denoted as $c_1 \neq c_2 \neq \cdots \neq c_n$. 

As illustrated in Figure~\ref{fig:method} (a), three draft models initiate sampling simultaneously when the queue $Q$ is initially empty.
In the subsequent stage, the draft models enter the queue according to which completes the generation first.
In Figure~\ref{fig:method} (a), Draft Model \symboldraftmodelone first completes the drafting process and is the first to enter the queue $Q$, followed by Draft Model \symboldraftmodeltwo and Draft Model \symboldraftmodelthree.
Each draft model is generating its own tokens while the target model $M_t$ is verifying the tokens of other draft models.
In this way, we can leverage the potential of small draft models to complete their drafting processes simultaneously, while the larger target model only needs to verify them sequentially.

\noindent \paragraph{Sequential Verification Phase}
Only one single target model is employed for the sequential verification of multiple draft sequences in \textsc{SeeD}.
The target model first verifies the tokens generated by the draft model at the front of the queue.
During the verification phase, two scenarios may occur: acceptance and rejection.
If the tokens generated by the draft model are accepted by the target model, they are retained, as exemplified by Draft Model \symboldraftmodelone in Fugure~\ref{fig:method} (a).
If rejected, one new token is resampled by the target model, as demonstrated by Draft Model \symboldraftmodeltwo and Draft Model \symboldraftmodelthree.
Taking Draft Model \symboldraftmodelthree as an example, it drafts two tokens, ``\textit{\textcolor{red}{many}}'' and ``\textit{\textcolor{red}{duch}}'', which are rejected by the target model.
Target Model \symboltargetmodel then resamples a new token ``\textit{\textcolor{blue}{much}}''.
Furthermore, when accepted, the target model only requires the execution time \symbolverify, when rejected, it incurs additional time for resampling \symbolresample.

\subsection{Rounds-Scheduled Strategy}
\label{sec:rounds_scheduled}
With the integration of parallel drafting and sequential verification, it is crucial to optimize the scheduling to ensure the correctness of speculative execution while effectively utilizing the target model and reducing the overall execution latency.

Inspired by process scheduling in operating system management, which utilizes the First-Come-First-Serve (FCFS) scheduling policy for all requests, ensuring fairness and preventing starvation~\citep{zhao1989performance, siahaan2016comparison}, we leverage a Rounds-Scheduled Strategy integrated with the FCFS scheduling policy to manage the verification process efficiently.
When a draft model completes its drafting phase and is ready for verification, the draft sequences along with $c$ are placed into a queue.

As depicted in Figure~\ref{fig:method} (a), when the queue $Q$ is not empty, a sequence of draft tokens is dequeued in the FCFS manner.
Target Model \symboltargetmodel first verifies the tokens generated by Draft Model \symboldraftmodelone, followed sequentially by tokens generated by Draft Model \symboldraftmodeltwo and Draft Model \symboldraftmodelthree, adhering to FCFS.
Upon completion of the verification of a draft sequence associated with a draft model, the draft model proceeds to the drafting process in the next iteration.

The overall scheduling diagram is shown in Figure~\ref{fig:method} (b), each draft model displays a series of iterations to complete the overall drafting progress for the Thought Generator or the State Evaluator.
The target model is consistently active across the overall scheduling timeline.
This continuous activity ensures that the target model is utilized efficiently, addressing issues related to idle time when acceptance rates are relatively high. 
Once all drafting and verification processes are completed, the entire execution concludes, resulting in the generation of $n$ sequences.

\begin{figure}[t]
    \centering
    \includegraphics[width=0.45\textwidth]{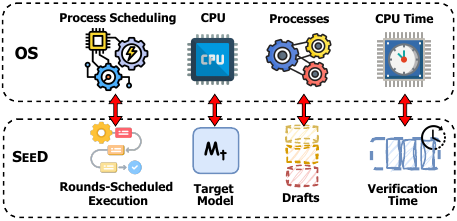}
    \caption{Analogy between the Operation System scheduler with our proposed \textsc{SeeD}.
    }
    \label{fig:analogy}
\end{figure}

\subsection{Technical Principle}\label{app:analogy}
Previous research has adapted the principle of the operating system (OS) scheduler for efficient process management~\citep{kwon2023efficient}.
As shown in Figure~\ref{fig:analogy}, each component in \textsc{SeeD} can be mapped to a corresponding component in the operating system scheduler.
We elaborate on each component individually as below:
\begin{itemize}[itemsep=0.5pt, parsep=0pt]
    \item The rounds-scheduled execution in \textsc{SeeD} corresponds to the process scheduling in OS.
    Both use an FCFS queue to control and maintain the overall execution flow.
    A key distinction exists: in \textsc{SeeD}, after the drafting tokens are processed by the verification phase, the draft model is returned to the queue, $\ie$, ``\textit{rounds}''. 
    In contrast, in OS scheduling, a process that has been handled by the CPU is marked as completed.
    \item The verification of draft tokens $\hat{\mathcal{X}}$ mirrors an operating process in OS scheduling.
    \item The target model serves $M_t$ analogously to the CPU.
    \item The total verification time of $M_t$ resembles the CPU time in OS process scheduling.
\end{itemize}

\subsection{Algorithm}
\label{sec:algorithm}
The core acceleration mechanisms of \textsc{SeeD}, which combines speculative scheduled execution with the rounds-scheduled strategy, is presented in Algorithm~\ref{alg:sse}.

\begin{algorithm}[t]
\caption{Speculative Scheduled Execution with a Rounds-Scheduled Strategy}
\small
\label{alg:sse}
\begin{algorithmic}[1]
\State \textbf{Input:} Draft models $\{M_{d_1},\cdots,M_{d_n}\}$, prefixes $\{c_1,\cdots,c_n\}$, target model $M_t$, max new length $l$, draft length $k$, 
auto-regressive drafting $p_{d_i}$ and length of current validated token $\mathcal{L}_i$ of the $i$-th draft model $M_{d_i}$,  \(i \in [1, n]\); 
\State \textbf{Initialize:} Prefill $\{M_{d_1},\cdots,M_{d_n}\}$ with prefixes; 
Create a verify queue $Q$ and a draft label map $\gamma$[i] of length $n$, with each element set to 1, \(i \in [1, n]\);
$\mathcal{L}_i \leftarrow 1$ , \(i \in [1, n]\);
Define $\hat{\mathcal{X}}_i[1:k]$ represents $\hat{x}_1, \dots, \hat{x}_{k}$ the sequence of draft tokens generated from $p_{d_i}$, \(i \in [1, n]\);
Start \(n\) draft processes ${\mathcal{D}}$(n) and 1 verification process $\mathcal{V}$ \textit{Synchronously}; 
\State \textbf{Processes ${\mathcal{D}}$(n):}  \textcolor{orange}{\Comment{Prallel Drafting}}
\While {$\exists i \in [1, n]: \mathcal{L}_i < l$}
    \If{$\gamma(i)$}
        \State $\hat{\mathcal{X}}_i[1:k] \leftarrow p_{d_i}(M_{d_i}, c_i, \hat{\mathcal{X}}_i[1:\mathcal{L}_i] ,k)$
        \State $Q \gets \hat{\mathcal{X}}_i[1:k]$ \textcolor{orange}{\Comment{Add draft tokens to the queue}}
        \State $\gamma[i] \gets 0$ \textcolor{orange}{\Comment{Draft Process D(i) wait}}
    \EndIf
\EndWhile
\State \textbf{Process $\mathcal{V}$:} \textcolor{blue}{\Comment{Sequential Verification}}
\While {$\exists i \in [1, n]: \mathcal{L}_i < l$}
    \If{Q is not empty}
        \State $\hat{\mathcal{X}}_i[1:k] \gets \text{queue}(Q)$ \textcolor{blue}{\Comment{FCFS}}
        \State $t_{1}, \cdots ,t_{k} \gets \mathcal{E}(M_t, c_i, \hat{\mathcal{X}}_i[1:k])$
        \For{$j=1$ \textbf{to} $k$} 
            \If {$t_j$ \text{ is acceptance}}
                \State $\hat{\mathcal{X}}_i[\mathcal{L}_i+1] \leftarrow \hat{x}_j$
                \State $\mathcal{L}_i\leftarrow \mathcal{L}_i+1$
            \Else
                \State $\hat{\mathcal{X}}[\mathcal{L}_i+1] \leftarrow \mathcal{R}(M_t, c_i, \hat{\mathcal{X}}_i[1: \mathcal{L}_i])$
                \State $\mathcal{L}_i\leftarrow \mathcal{L}_i+1$
                \State Break
            \EndIf
        \EndFor
        \State $\gamma[i] \gets 1$ \textcolor{blue}{\Comment{Draft Process D(i) continue}}
    \EndIf
\EndWhile
\State Wait for all $\mathcal{D}$(n) and $\mathcal{V}$ to finish
\State \Return $[response_1, \dots, response_n]$
\end{algorithmic}
\end{algorithm}

At its essence, the parallel drafting is realized by multiple parallel processes $\mathcal{D}(n)$, while the sequential verification is realized by a verification process $\mathcal{V}$ that cyclically verifies from the verify queue $\mathcal{Q}$.
The verification process has two phases, which are the verify phase $\mathcal{E}$ and the resampling phase $\mathcal{R}$.
To maintain the asynchronous nature of the \textit{draft-then-verify} event loop, leveraging a draft label map $\gamma$ ensures each draft process waits for verification before proceeding with new drafts.
At the initial stage, each element in the draft label map $\gamma$ is set to 1, indicating all draft models can perform drafting.
After completing the verification of a draft model, the corresponding label in $\gamma$ changes to 0, awaiting for re-drafting.
Notably, $\mathcal{D}(n)$ and $\mathcal{V}$ are \textit{synchronized}.
The termination condition for both process $\mathcal{D}(n)$ and process $\mathcal{V}$ is that all current validated token $\mathcal{L}_i, i \in [1, n]$ equals the max new length $l$.
When all the processes are finished, we can obtain a list containing $n$ response.

\section{Experiments}

\subsection{Datasets}
Three widely used reasoning and planning datasets are chosen for our experiments.
To assess the effectiveness of creativity and planning tasks, we leverage the Creative Writing dataset (CW)~\citep{yao2024tree}, where the input is four random sentences and the output should be a coherent passage with four paragraphs that end in the four input sentences respectively, with a ToT tree depth $\mathcal{T}$ of 2.
For mathematical reasoning, 
GSM8K~\citep{cobbe2021training} is a dataset comprising high-quality grade-school math word problems that require multi-step reasoning, with a tree depth $\mathcal{T}$ of 4.
This task is open-ended and exploratory, posing significant challenges to creative thinking and high-level planning.
To better demonstrate the speedup performance in solving more complex planning problems, we select the Blocksworld dataset (BW)~\citep{valmeekam2023planning}.
We set the tree depth $\mathcal{T}$ to 7 for this task to allow for more iterations.
Specifically, we utilize 1319 samples from the GSM8K test set, 100 random samples from the CW dataset following~\cite{yao2024tree}, and 145 samples from the BW step-6 dataset.

\begin{table*}[t]
    \small
    \centering
    \begin{adjustbox}{width=0.92\textwidth}
    \begin{tabular}{c|c|c|cccccc}
    \toprule
    \multirow{2}*{\textbf{Temp.}} & \multirow{2}*{\textbf{$k_{\text{config}}$}} & \multirow{2}*{\textbf{Methods}} & \multicolumn{2}{c}{\textbf{CW($\mathcal{T}=2$)}} & \multicolumn{2}{c}{\textbf{GSM8K($\mathcal{T}=4$)}} & \multicolumn{2}{c}{\textbf{BW($\mathcal{T}=7$)}}\\
    & & & Tokens/s & Speedup & Tokens/s & Speedup & Tokens/s & Speedup\\
    \midrule \midrule
    \multirow{11}{*}{\textbf{0.2}}
    & - & AR & 38.42 & 1.000$\times$ & 42.31 & 1.000$\times$ & 34.19 & 1.000$\times$\\
    \cmidrule{2-9}
    & \multirow{5}{*}{(1,1,1)} & SD & 39.96 & 1.040$\times$ & 51.11 & 1.208$\times$ & 36.28 & 1.061$\times$\\    
    &  & \cellcolor{mycell}{\textbf{w. \textsc{SeeD}}} & \cellcolor{mycell}{\textbf{41.53}} & \cellcolor{mycell}{\textbf{1.081$\times$}} & \cellcolor{mycell}{\textbf{53.14}} & \cellcolor{mycell}{\textbf{1.256$\times$}} & \cellcolor{mycell}{\textbf{36.93}} & \cellcolor{mycell}{\textbf{1.080$\times$}}\\
    \cmidrule{3-9}
    &  & MCSD & 40.19 & 1.046$\times$ & 52.42 & 1.239$\times$ & 36.04 & 1.054$\times$\\    
    &  & \cellcolor{mycell}{\textbf{w. \textsc{SeeD}}} & \cellcolor{mycell}{\textbf{41.46}} & \cellcolor{mycell}{\textbf{1.079$\times$}} & \cellcolor{mycell}{\textbf{53.78}} & \cellcolor{mycell}{\textbf{1.271$\times$}} & \cellcolor{mycell}{\textbf{36.96}} & \cellcolor{mycell}{\textbf{1.081$\times$}}\\
    \cmidrule{2-9}
    & \multirow{5}{*}{(2,2,1)} & SD & 46.22 & 1.203$\times$ & 60.63 & 1.433$\times$ & 40.04 & 1.171$\times$\\    
    &  & \cellcolor{mycell}{\textbf{w. \textsc{SeeD}}} & \cellcolor{mycell}{\textbf{48.60}} & \cellcolor{mycell}{\textbf{1.265$\times$}} & \cellcolor{mycell}{\textbf{65.24}} & \cellcolor{mycell}{\textbf{1.542$\times$}} & \cellcolor{mycell}{\textbf{44.24}} & \cellcolor{mycell}{\textbf{1.294$\times$}}\\
    \cmidrule{3-9}
    &  & MCSD & 46.80 & 1.218$\times$ & 60.88 & 1.439$\times$ & 40.79 & 1.193$\times$\\    
    &  & \cellcolor{mycell}{\textbf{w. \textsc{SeeD}}} & \cellcolor{mycell}{\textbf{48.79}} & \cellcolor{mycell}{\textbf{1.270$\times$}} & \cellcolor{mycell}{\textbf{65.58}} & \cellcolor{mycell}{\textbf{1.550$\times$}} & \cellcolor{mycell}{\textbf{44.75}} & \cellcolor{mycell}{\textbf{1.309$\times$}}\\
    \midrule
    \multirow{11}{*}{\textbf{1.0}}
    & - & AR & 39.47 & 1.000$\times$ & 47.81 & 1.000$\times$ & 34.62 & 1.000$\times$\\
    \cmidrule{2-9}
    & \multirow{5}{*}{(1,1,1)} & SD & 45.90 & 1.163$\times$ & 55.32 & 1.157$\times$ & 35.14 & 1.015$\times$\\    
    &  & \cellcolor{mycell}{\textbf{w. \textsc{SeeD}}} & \cellcolor{mycell}{\textbf{46.77}} & \cellcolor{mycell}{\textbf{1.185$\times$}} & \cellcolor{mycell}{\textbf{61.01}} & \cellcolor{mycell}{\textbf{1.276$\times$}} & \cellcolor{mycell}{\textbf{38.94}} & \cellcolor{mycell}{\textbf{1.125$\times$}}\\
    \cmidrule{3-9}
    &  & MCSD & 45.63 & 1.156$\times$ & 58.47 & 1.223$\times$ & 38.05 & 1.099$\times$\\    
    &  & \cellcolor{mycell}{\textbf{w. \textsc{SeeD}}} & \cellcolor{mycell}{\textbf{46.54}} & \cellcolor{mycell}{\textbf{1.179$\times$}} & \cellcolor{mycell}{\textbf{65.50}} & \cellcolor{mycell}{\textbf{1.370$\times$}} & \cellcolor{mycell}{\textbf{40.02}} & \cellcolor{mycell}{\textbf{1.156$\times$}}\\
    \cmidrule{2-9}
    & \multirow{5}{*}{(2,2,1)} & SD & 57.39 & 1.454$\times$ & 66.74 & 1.396$\times$ & 45.98 & 1.328$\times$\\    
    &  & \cellcolor{mycell}{\textbf{w. \textsc{SeeD}}} & \cellcolor{mycell}{\textbf{58.89}} & \cellcolor{mycell}{\textbf{1.492$\times$}} & \cellcolor{mycell}{\textbf{72.62}} & \cellcolor{mycell}{\textbf{1.519$\times$}} & \cellcolor{mycell}{\textbf{47.22}} & \cellcolor{mycell}{\textbf{1.364$\times$}}\\
    \cmidrule{3-9}
    &  & MCSD & 56.24 & 1.425$\times$ & 67.36 & 1.409$\times$ & 46.18 & 1.334$\times$\\    
    &  & \cellcolor{mycell}{\textbf{w. \textsc{SeeD}}} & \cellcolor{mycell}{\textbf{59.76}} & \cellcolor{mycell}{\textbf{1.514$\times$}} & \cellcolor{mycell}{\textbf{74.44}} & \cellcolor{mycell}{\textbf{1.557$\times$}} & \cellcolor{mycell}{\textbf{47.71}} & \cellcolor{mycell}{\textbf{1.378$\times$}}\\
    \midrule
    \bottomrule
  \end{tabular}
  \end{adjustbox}
\caption{The speedup performance of our proposed \textsc{SeeD} and baselines, with settings of \textsc{SeeD} for $M_d$ and $M_t$ being LLaMA-68M and LLaMA2-7B, respectively.
The illustration of $k_{\text{config}}$=(2,2,1) is presented in Appendix~\ref{app:tree_attention}.
All speedups are relative to the vanilla AR.
The best results among all methods are in \textbf{bolded}.}
\label{table:result_gsm8k_writing_blocksworld_7b}
\end{table*}

\subsection{Baselines}
\label{sec:baseline}
This study focuses on accelerating the reasoning tree construction process rather than the search algorithm or advanced prompting methods.
The selection of baselines will be discussed in Appendix~\ref{app:baseline}.
We consider the following decoding paradigms as our baselines:
(1) \textbf{AR} denotes the original ToT~\cite{yao2024tree} that employing standard autoregressive generation as shown in Figure~\ref{fig:intro} (a);
(2) \textbf{SD} presents the application of speculative sampling which is detailed in~\ref{sec:preliminaries} on the basis of ToT as shown in Figure~\ref{fig:intro} (b);
(3) \textbf{MCSD} utilizes multi-candidate sampling and employs a advanced verifying algorithm to improve the acceptance rate and enhance the speed of SD~\citep{yang2024multi}.
Similar to SD, it adheres to only one single-sample serial execution process.
Notably, both SD and MCSD are \textbf{orthogonal} to our proposed \textsc{SeeD}.
We apply our framework within these two decoding approaches to validate \textsc{SeeD}'s effectiveness across different accetance rates. 

\subsection{Setup}
\noindent \paragraph{Model Suite}
Our evaluation is based on the publicly available LLaMA Chat suite~\citep{touvron2023llama2}, which has shown strong performance in executing instructions and in TSB scenarios.
We utilize ($M_{d}$, $M_{t}$) following previous work~\citep{chen2023cascade, yang2024multi}: (LLaMA-68M-Chat\footnote{{\scriptsize \url{https://huggingface.co/Felladrin/Llama-68M-Chat-v1}}} , LLaMA-2-Chat-7B\footnote{{\scriptsize \url{https://huggingface.co/meta-llama/Llama-2-7b-chat-hf}}}) and (LLaMA-160M-Chat\footnote{{\scriptsize \url{https://huggingface.co/Felladrin/Llama-160M-Chat-v1}}} , LLaMA-2-Chat-13B\footnote{{\scriptsize \url{https://huggingface.co/meta-llama/Llama-2-13b-chat-hf}}}).
To validate the extensibility of our framework, we also conducted experiments using the QWen suite~\citep{bai2023qwen}.
Detailed information and results for both the other LLaMA pair and QWen suite can be found in Appendix~\ref{app:model_extensibility}.

\begin{figure*}[t]
    \centering
    \includegraphics[width=0.85\textwidth]{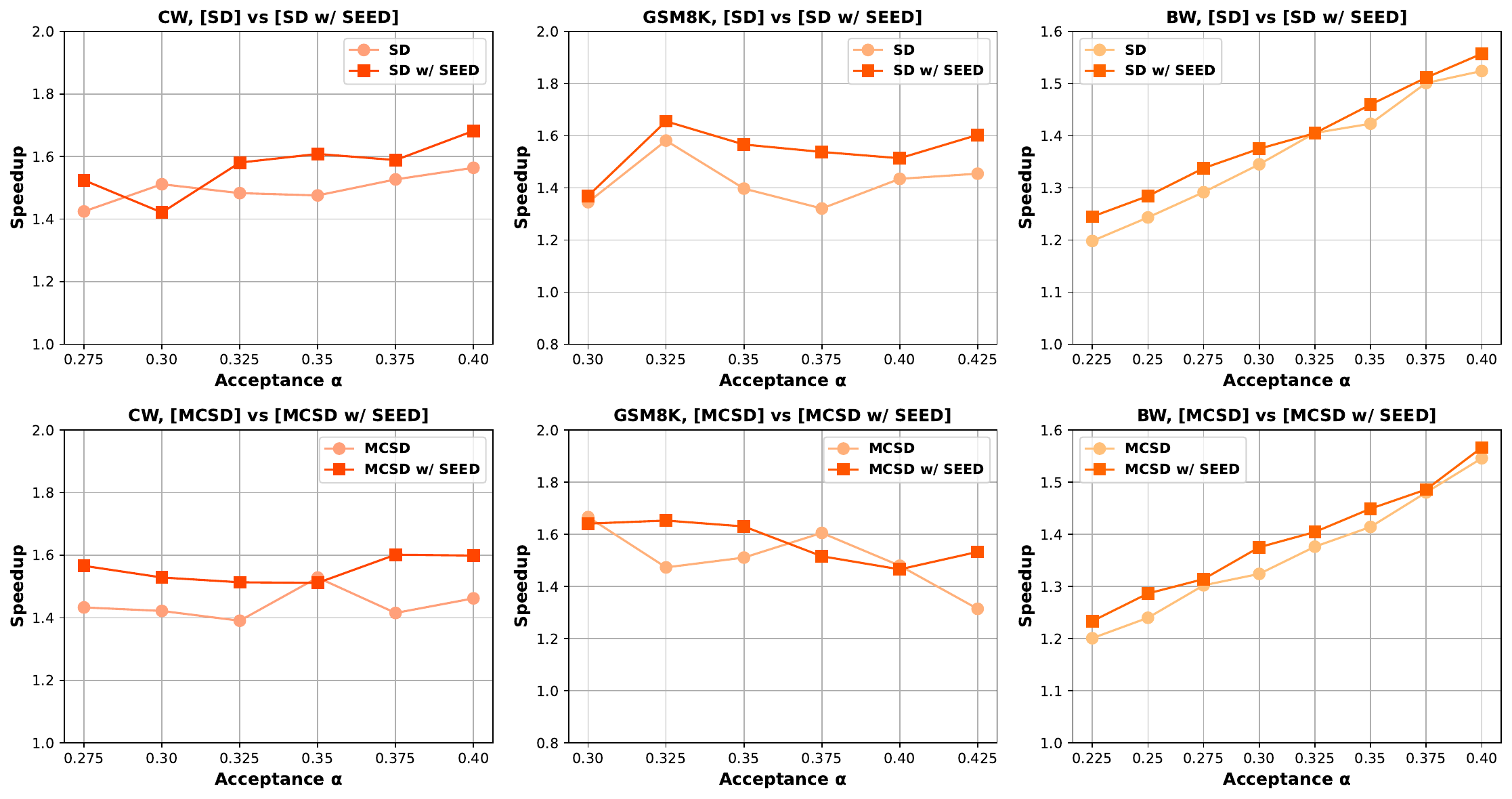}
    \caption{The variation of speedup performance across three datasets at different acceptance rates $\alpha$.}
    \label{fig:acc_speedup}
\end{figure*}

\noindent \paragraph{Hyperparameters}
We perform a BFS algorithm as the search strategy.
Temperatures are set to 0.2 and 1.0 to evaluate under different conditions.\footnote{We avoid the  temperature 0 because greedy decoding is not meaningful in Thought Generator.}
The detailed prompts for the Thought Generator and the State Evaluator, along with the ToT setup for each task are provided in Appendix~\ref{app:prompts}.

\noindent \paragraph{Environment}
The experiments are conducted on a single NVIDIA RTX A100-80G or a single node which is equipped with four
NVIDIA RTX 3090-24GB GPUs.
Subtle differences in hardware performance between these platforms are discussed in Appendix~\ref{app:hardware}.

\section{Results and Analysis}
\subsection{Main Results}
\label{sec:main_results}
Table~\ref{table:result_gsm8k_writing_blocksworld_7b} presents a comprehensive analysis of our proposed \textsc{SeeD} and baselines applied to three reasoning datasets. 
If each element in $k_{\text{config}}$ is 1, we use the traditional single sampling at each position of the draft sequence.
Otherwise, we employ tree attention, which represents sample multiple candidate tokens at each position and verify in parallel (details in Section~\ref{sec:preliminaries}).
A greater number at each position in $k_{\text{config}}$ signifies that more candidates, generally yield higher speedups.
MCSD achieves better speedup than SD by using an advanced verifying algorithm that results in higher acceptance rates.
With our \textsc{SeeD}, the performance of these two baselines is further improved, demonstrating its effectiveness across different acceptance rates.
Across all datasets across various reasoning depths $\mathcal{T}$, our framework, consistently outperforms the baselines across different settings and configurations, including temperature and $k_\text{config}$, in terms of speedup, achieving the further speedup. 
Specifically, on the GSM8K dataset, using tree attention, MCSD in our proposed \textsc{SeeD} framework achieves up to $1.5\times$ speedup compared to AR, generating nearly 30 additional tokens per second.

\begin{figure}[t]
    \centering
    \includegraphics[width=0.48\textwidth]{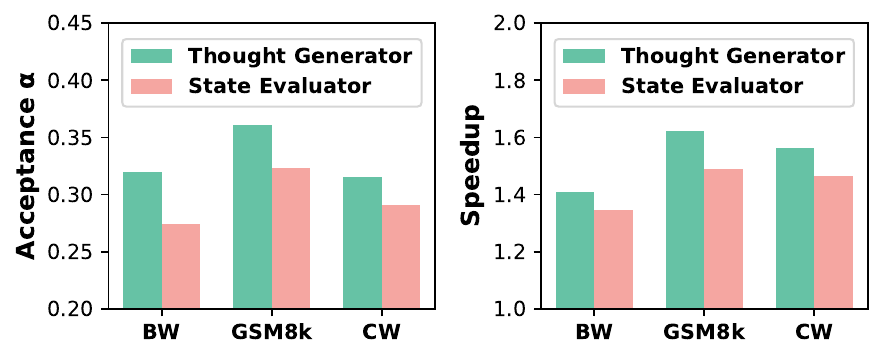}
    \caption{The acceptance rate $\alpha$ and the speedup performance of the Thought Generator and the State Evaluator.}
    \label{fig:two_components_acc_speedup}
\end{figure}

\subsection{Analysis}
We use the \textsc{SeeD} (with MCSD) to conduct the following analytical experiment to answer the following research question (\emph{RQ}) using under the condition $k_{\text{config}}$ = (2,2,1) and temperature = 1.0.

\noindent \paragraph{\emph{RQ1: How does \textsc{SeeD} perform at different acceptance rates?}}
We sampled data points from three datasets within different acceptance rate ranges, we separately reported the speedup achieved by \textsc{SeeD} and the baseline for these samples in Figure~\ref{fig:acc_speedup}.
It is evident that under the same acceptance rate, \textsc{SeeD} outperforms the baseline in terms of speedup.
This improvement is attributed to our framework, which achieves speedup not by increasing the acceptance rate but by scheduling draft models. 
Additionally, as the acceptance rate increases, both \textsc{SeeD} and the baseline exhibit a noticeable upward trend in speedup, which is the inherent characteristic of the speculative decoding method.
 
\begin{figure}[t]
    \centering
    \includegraphics[width=0.49\textwidth]{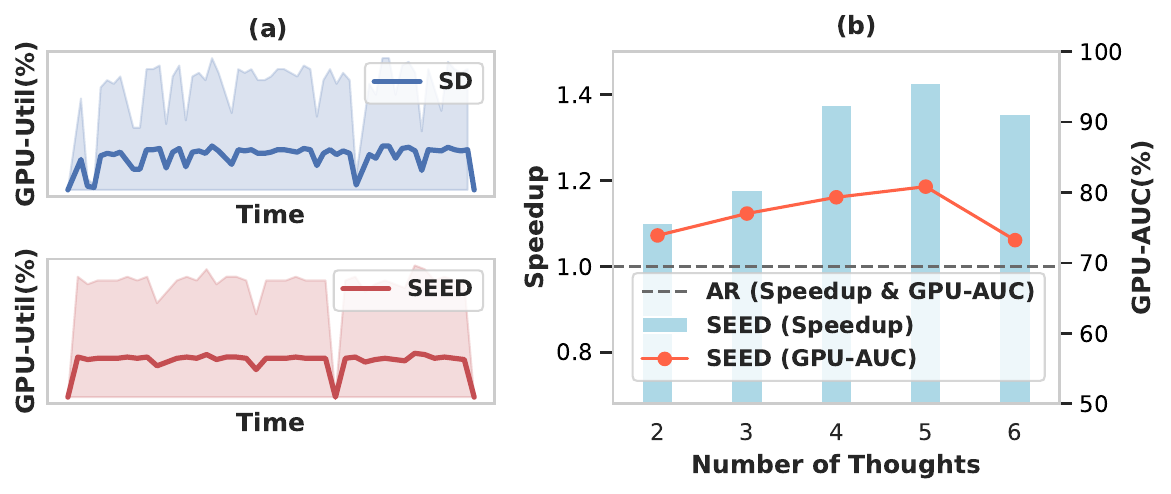}
    \caption{(a)  The comparison visualization of GPU utilization between SD and \textsc{SeeD} over the 120 seconds under $n$ = 3.
    (b) The variation of speedup and acceptance rate $\alpha$ with the number of reasoning paths $n$.}
    \label{fig:tht_num_speedup_gpu_auc}
\end{figure}

\noindent \paragraph{\emph{RQ2: Does \textsc{SeeD} exhibit different acceleration effects on different components of ToT?}}
\textsc{SeeD} accelerate two components in reasoning tree construction, which are the TG and the SE.
Figure~\ref{fig:two_components_acc_speedup} presents the acceptance rate $\alpha$ and the speedup performance of two main components of the \textsc{SeeD} method on the GSM8K dataset, confirming that the answer to the \emph{RQ2} is \textbf{Yes}.
The TG executes multiple iterations with the same prompt while the SE refers to evaluates multiple iterations with different prompts.
The TG component consistently outperforms the SE component in terms of both $\alpha$ and speedup, possibly because the SE is relatively harder compared to the TG.
The proficiency between the target model and draft model may be more closely aligned in the proposal of thoughts, compared to decision-making capability.
\noindent \paragraph{\emph{RQ3: How does the speedup and GPU utilization scale with the number of thoughts?}}
In speculative decoding, both the target and draft model parameters are loaded into GPU memory.
We record the GPU utilization over the same durations for the SD and \textsc{SeeD} on a GSM8K instance to visualize the effectiveness of parallel drafting in Figure~\ref{fig:tht_num_speedup_gpu_auc} (a).
The upper part illustrates the GPU utilization of SD fluctuates intermittently, primarily due to the target model being idle during drafting, while the lower part shows \textsc{SeeD} exhibits stable utilization, attributed to the active engagement of the target model in the verification phase.
As the number of thoughts $n$ increases within a certain range, the idle time of the target model decreases, leading to higher GPU utilization and speedup, as shown in Figure~\ref{fig:tht_num_speedup_gpu_auc} (b).
However, when the number of thoughts becomes too large (\eg, $n$=6), the target model's fixed verification capacity leads to \textsc{SeeD} speedup saturation. 
This manifests as more draft models being placed in a waiting state, reducing draft parallelism and causing bottlenecks that lower utilization and acceleration.

\section{Conclusion and Discussion}
In this paper, we introduce \textsc{SeeD}, a novel inference framework designed to optimize the runtime speed and manage GPU memory usage effectively during the reasoning tree construction for complex reasoning and planning tasks.
\textsc{SeeD} employs scheduled speculative execution to enhance the performance of LLMs by integrating the management of multiple draft models and a single target model, based on principles similar to operating system process scheduling.
This strategy not only mitigates the inference latency inherent in tree-search-based reasoning methods but also efficiently utilizes the available computational resources.
Our extensive experimental evaluation across three reasoning demonstrates that \textsc{SeeD} achieves significant improvements in inference speed, generating up to 20 additional tokens per second.

Our work accelerates the development of ToT, offering the potential for \textbf{seamless} extension to further advance the test-time scaling of LLMs~\citep{zhang2024llama, xie2024monte, zhang2024chain,snell2024scaling,wu2024comparative}.
This framework and vision represent a promising direction for improving the efficiency of LLM reasoning in real-world applications.

\section*{Limitations}
Although \textsc{SeeD} already achieves exceptional speedup performance in the experiments, our work also has the following limitations.

\begin{itemize}[itemsep=0.5pt, parsep=0pt]
\item Our frameworks introduce parallel drafting, involving $n-1$ additional drafting models, which inherently necessitates the addition of an equivalent number of KV-Cache.
Given the increase attributed to small draft models (68M/160M) is relatively minimal, we do not optimize the management of the KV-Cache in this work.

\item This study focuses solely on optimizing the inference speed of the tree construction for the TSB reasoning task and does not optimize the search speed for these tasks.
Our framework uses the relatively simple search algorithm BFS.
In fact, \textsc{SeeD} can seamlessly integrate more advanced search algorithms, such as $A^*$~\citep{hart1968formal} and MCTS~\citep{kocsis2006bandit}, $\etc$, which we leave for future research.
\item We employ the most widely used scheduling algorithm, FCFS.
Future work may explore the integration of more advanced scheduling algorithms, such as those used in real-time systems, to further enhance the responsiveness and efficiency of \textsc{SeeD}.
\end{itemize}

In the future, \textsc{SeeD} can be compatible with vLLM~\citep{kwon2023efficient} and FlashAttention-2~\citep{dao2024flashattention}, enabling more memory-efficient inference on longer sequences.
Additionally, the extra KV-Cache could be reduced by caching the common prefix during reasoning tree construction, which would lower the parallel overhead in later iterations.

Moreover, our method offers a potential implementation of batched speculative decoding from the execution scheduling perspective, which could be integrated with other KV-Cache based batch speculative decoding methods~\cite{ni2024ems}, as further discussed in Appendix~\ref{app:batch_infer}.

\section*{Acknowledgement}
The authors would like to thank the anonymous reviewers for their insightful comments. This work is funded by the National Natural Science Foundation of China (Grant No.62176053). This work is supported by the Big Data Computing Center of Southeast University.

\bibliography{custom}
\clearpage
\newpage
\appendix

\section{Discussions}
\subsection{Selection of Baselines}\label{app:baseline}
See Section~\ref{sec:baseline}, where we list all the baselines used to compare with our proposed \textsc{SeeD} in this study. 
However, several other speculative decoding strategies have not been explored as baselines.
We do not conclude these strategies based on the following considerations as shown in Table~\ref{tab:baseline_comparison}:

\textbf{(1) Training-free} indicates whether the method requires training. 
\begin{itemize}[itemsep=0.5pt, parsep=0pt,label=$\ast$]
    \item \textbf{Medusa}~\citep{cai2024medusa} adds extra FFN heads atop the Transformer decoder, allowing for parallel token generation at each step;
    \item \textbf{Eagle}~\citep{li2024eagle} performs the drafting process autoregressively at a more structured level, specifically the second-to-top layer of features;
    \item \textbf{SS}~\citep{bhendawade2024speculative} integrates drafting phase into the target model by modifying the fine-tuning objective from the next token to future n-gram predictions.
\end{itemize}
These methods all \textbf{require training and are not plug-and-play}, since they train the LLM to serve as both the target model and the draft model, which classifies them as self-drafting~\squareSymbol~according to \citet{xia-etal-2024-unlocking};
in contrast, our method employs independent drafting~\triangleSymbol~(draft-and-target), placing it in a different SD type.
Therefore, we do not consider them as baselines.

\textbf{(2) Extra-knowledge-free} indicates whether the SD process uses additional knowledge modules.
\begin{itemize}[itemsep=0.5pt, parsep=0pt,label=$\ast$]
    \item \textbf{CS-drafting}~\citep{chen2023cascade} resorts to a bigram model based on the probability distribution of Wikipedia as the draft model at a more basic level.
    \item \textbf{REST}~\citep{he2023rest} retrieve from extensive code and conversation data stores to generate draft tokens.
\end{itemize}
The two approaches introduce external knowledge modules, making it significantly dependent on the effectiveness of the external knowledge modules and unfair to compare us with draft-and-target models.

\textbf{(3) Lossless} indicates whether the method generates the same output distribution as AR decoding does in the backbone model.

\textbf{SS}~\citep{bhendawade2024speculative} and \textbf{Medusa}~\citep{cai2024medusa}, which are inherently not lossless, are unsuitable for comparison with \textbf{our proposed \textsc{SeeD}, which maintains losslessness consistent with SD in a single \textit{draft-then-verify}}.

\begin{table}[t]
    \centering
    \small
    \resizebox{\columnwidth}{!}{%
    \begin{tabular}{c|c|c|c|c}
    \toprule
    \textbf{Temp.} & $k_{\text{config}}$ & \textbf{Methods}  & \textbf{Tokens/s} & \textbf{Speedup} \\
    \midrule
    \multirow{3}{*}{\textbf{0.2}} & - 
    & AR & 31.22 & 1.000$\times$ \\
    \cmidrule{2-5}
    & \multirow{2}{*}{(1,1,1,1)} & SD & 32.91 & 1.054$\times$ \\
    & & \cellcolor{mycell}{\textbf{w. \textsc{SeeD}}} & \cellcolor{mycell}{\textbf{34.62}} & \cellcolor{mycell}{\textbf{1.109$\times$}} \\
    \midrule
    \multirow{3}{*}{\textbf{0.6}} & -  
    & AR & 37.93 & 1.000$\times$ \\
    \cmidrule{2-5}
    & \multirow{2}{*}{(1,1,1,1)} & SD & 39.22 & 1.034$\times$ \\
    & & \cellcolor{mycell}{\textbf{w. \textsc{SeeD}}} & \cellcolor{mycell}{\textbf{41.91}} & \cellcolor{mycell}{\textbf{1.105$\times$}} \\
    \midrule
    \multirow{3}{*}{\textbf{1}} & - 
    & AR & 33.86 & 1.000$\times$ \\
    \cmidrule{2-5}
    & \multirow{2}{*}{(1,1,1,1)} & SD & 34.91 & 1.031$\times$ \\
    & & \cellcolor{mycell}{\textbf{w. \textsc{SeeD}}} & \cellcolor{mycell}{\textbf{39.35}} & \cellcolor{mycell}{\textbf{1.162$\times$}} \\
    \bottomrule
  \end{tabular}
  }
\caption{Speedup performance on Creative Writing dataset of \textsc{SeeD} within using QWen1.5-0.5B-Chat as $M_d$ and QWen1.5-7B-Chat as $M_t$. The vocabularies of these two models are identical, allowing for speculative sampling.}
\label{table:qwen1.5_speedup}
\end{table}

\subsection{Scalability and Extensibility}\label{app:model_extensibility}
\noindent \paragraph{LLaMA Suite} Table~\ref{table:result_gsm8k_writing_blocksworld_13b} shows the performance of each method when using LLaMA-160M-Chat\footnote{{\scriptsize \url{https://huggingface.co/Felladrin/Llama-160M-Chat-v1}}} as draft model $M_d$ and LLaMA-2-Chat-13B\footnote{{\scriptsize \url{https://huggingface.co/meta-llama/Llama-2-13b-chat-hf}}} as target model $M_t$.

\begin{table*}[t]
    \small
    \centering
    \begin{tabular}{c|c|c|cccccc}
    \toprule
    \multirow{2}*{\textbf{Temp.}} & \multirow{2}*{\textbf{$k_{\text{config}}$}} & \multirow{2}*{\textbf{Methods}} & \multicolumn{2}{c}{\textbf{CW($\mathcal{T}=2$)}} & \multicolumn{2}{c}{\textbf{GSM8K($\mathcal{T}=4$)}} & \multicolumn{2}{c}{\textbf{BW($\mathcal{T}=7$)}}\\
    & & & Tokens/s & Speedup & Tokens/s & Speedup & Tokens/s & Speedup\\
    \midrule \midrule
    \multirow{11}{*}{\textbf{0.2}}
    & - & AR & 32.33 & 1.000$\times$ & 32.08 & 1.000$\times$ & 32.91 & 1.000$\times$\\
    \cmidrule{2-9}
    & \multirow{5}{*}{(2,1,1)} & SD & 33.14 & 1.025$\times$ & 34.97 & 1.090$\times$ & 33.17 & 1.008$\times$\\    
    &  & \cellcolor{mycell}{\textbf{w. \textsc{SeeD}}} & \cellcolor{mycell}{\textbf{33.82}} & \cellcolor{mycell}{\textbf{1.046$\times$}} & \cellcolor{mycell}{\textbf{36.80}} & \cellcolor{mycell}{\textbf{1.147$\times$}} & \cellcolor{mycell}{\textbf{33.54}} & \cellcolor{mycell}{\textbf{1.019$\times$}}\\
    \cmidrule{3-9}
    &  & MCSD & 33.27 & 1.029$\times$ & 35.71 & 1.113$\times$ & 33.37 & 1.014$\times$\\    
    &  & \cellcolor{mycell}{\textbf{w. \textsc{SeeD}}} & \cellcolor{mycell}{\textbf{36.18}} & \cellcolor{mycell}{\textbf{1.119$\times$}} & \cellcolor{mycell}{\textbf{36.28}} & \cellcolor{mycell}{\textbf{1.131$\times$}} & \cellcolor{mycell}{\textbf{34.36}} & \cellcolor{mycell}{\textbf{1.044$\times$}}\\
    \cmidrule{2-9}
    & \multirow{5}{*}{(4,2,1)} & SD & 34.23 & 1.059$\times$ & 38.95 & 1.214$\times$ & 36.04 & 1.095$\times$\\    
    &  & \cellcolor{mycell}{\textbf{w. \textsc{SeeD}}} & \cellcolor{mycell}{\textbf{38.57}} & \cellcolor{mycell}{\textbf{1.193$\times$}} & \cellcolor{mycell}{\textbf{41.06}} & \cellcolor{mycell}{\textbf{1.280$\times$}} & \cellcolor{mycell}{\textbf{36.76}} & \cellcolor{mycell}{\textbf{1.117$\times$}}\\
    \cmidrule{3-9}
    &  & MCSD & 35.56 & 1.100$\times$ & 41.09 & 1.281$\times$ & 37.58 & 1.142$\times$\\    
    &  & \cellcolor{mycell}{\textbf{w. \textsc{SeeD}}} & \cellcolor{mycell}{\textbf{40.28}} & \cellcolor{mycell}{\textbf{1.246$\times$}} & \cellcolor{mycell}{\textbf{44.11}} & \cellcolor{mycell}{\textbf{1.375$\times$}} & \cellcolor{mycell}{\textbf{38.70}} & \cellcolor{mycell}{\textbf{1.176$\times$}}\\
    \midrule
    \multirow{11}{*}{\textbf{1.0}}
    & - & AR & 39.57 & 1.000$\times$ & 31.54 & 1.000$\times$ & 32.87 & 1.000$\times$\\
    \cmidrule{2-9}
    & \multirow{5}{*}{(2,1,1)} & SD & 40.28 & 1.018$\times$ & 35.23 & 1.117$\times$ & 34.32 & 1.044$\times$\\    
    &  & \cellcolor{mycell}{\textbf{w. \textsc{SeeD}}} & \cellcolor{mycell}{\textbf{42.74}} & \cellcolor{mycell}{\textbf{1.080$\times$}} & \cellcolor{mycell}{\textbf{36.71}} & \cellcolor{mycell}{\textbf{1.164$\times$}} & \cellcolor{mycell}{\textbf{35.37}} & \cellcolor{mycell}{\textbf{1.076$\times$}}\\
    \cmidrule{3-9}
    &  & MCSD & 40.68 & 1.028$\times$ & 35.26 & 1.118$\times$ & 35.01 & 1.065$\times$\\    
    &  & \cellcolor{mycell}{\textbf{w. \textsc{SeeD}}} & \cellcolor{mycell}{\textbf{43.37}} & \cellcolor{mycell}{\textbf{1.096$\times$}} & \cellcolor{mycell}{\textbf{37.15}} & \cellcolor{mycell}{\textbf{1.178$\times$}} & \cellcolor{mycell}{\textbf{35.86}} & \cellcolor{mycell}{\textbf{1.091$\times$}}\\
    \cmidrule{2-9}
    & \multirow{5}{*}{(4,2,1)} & SD & 43.69 & 1.104$\times$ & 36.87 & 1.169$\times$ & 37.83 & 1.151$\times$\\    
    &  & \cellcolor{mycell}{\textbf{w. \textsc{SeeD}}} & \cellcolor{mycell}{\textbf{47.25}} & \cellcolor{mycell}{\textbf{1.194$\times$}} & \cellcolor{mycell}{\textbf{40.66}} & \cellcolor{mycell}{\textbf{1.289$\times$}} & \cellcolor{mycell}{\textbf{38.56}} & \cellcolor{mycell}{\textbf{1.173$\times$}}\\
    \cmidrule{3-9}
    &  & MCSD & 45.19 & 1.142$\times$ & 36.90 & 1.170$\times$ & 39.28 & 1.195$\times$\\    
    &  & \cellcolor{mycell}{\textbf{w. \textsc{SeeD}}} & \cellcolor{mycell}{\textbf{49.74}} & \cellcolor{mycell}{\textbf{1.257$\times$}} & \cellcolor{mycell}{\textbf{41.54}} & \cellcolor{mycell}{\textbf{1.317$\times$}} & \cellcolor{mycell}{\textbf{40.43}} & \cellcolor{mycell}{\textbf{1.230$\times$}}\\
    \midrule
    \bottomrule
  \end{tabular}
\caption{Speedup performance of our proposed \textsc{SeeD} and baselines, with settings of \textsc{SeeD} for $M_d$ and $M_t$ being LLaMA-160M and LLaMA2-13B, respectively.
All speedups are relative to the vanilla AR.
The best results among all methods are in \textbf{bolded}.}
\label{table:result_gsm8k_writing_blocksworld_13b}
\end{table*}

\noindent \paragraph{QWen Suite}
Our framework is based on speculative decoding, so the model setup of the draft model and the target model can be consistent with it.
Consequently, any LLM suite can be integrated into our framework.
We also conducted experiments using the QWen1.5 suite.\footnote{\scriptsize \url{https://qwenlm.github.io/zh/blog/qwen1.5/}}
Specifically, we use QWen1.5-0.5B-Chat\footnote{\scriptsize \url{https://huggingface.co/Qwen/Qwen1.5-0.5B-Chat}} as the draft model $M_d$ and use QWen1.5-7B-Chat\footnote{\scriptsize \url{https://huggingface.co/Qwen/Qwen1.5-7B-Chat}} as the target model $M_t$.
The results are presented in Table~\ref{table:qwen1.5_speedup}.
The results align with the findings presented in Section~\ref{sec:main_results}, demonstrating the superior performance of our framework.
It also highlights the scalability of our framework to the LLM suite~\citep{bai2023qwen}.

\begin{table*}[t]
\centering
    \small
    \begin{tabular}{lcccccc}
    \toprule
    \textbf{Methods} & \textbf{Training-free}  &  \textbf{Lossless} & \textbf{SD Type} & \textbf{Extra-knowledge-free} & \textbf{Speedup} \\
    \midrule
    Vanilla AR  & \yes & \yes & - & \yes & \no \\
    \midrule
    SD~\cite{leviathan2023fast} & \yes & \yes & \triangleSymbol & \yes & \yes \\
    \midrule
    CS-Drafting~\cite{chen2023cascade} & \yes & \yes & \triangleSymbol & \no & \yes \\
    \midrule
    REST~\cite{he2023rest} & \yes & \yes & \triangleSymbol & \no & \yes \\
    \midrule
    Medusa ~\cite{cai2024medusa} & \no & \no & \squareSymbol & \yes & \yes \\
    \midrule
    Eagle~\cite{li2024eagle} & \no & \yes & \squareSymbol & \yes & \yes \\
    \midrule
    SS~\cite{bhendawade2024speculative} & \no & \no & \squareSymbol & \yes & \yes \\
    \midrule
    MCSD~\cite{yang2024multi} & \yes & \yes & \triangleSymbol & \yes & \yes \\
    \midrule
    \textbf{\textsc{SeeD} (Ours)} & \yes & \yes & \triangleSymbol & \yes & \yes \\
    \bottomrule
    \end{tabular}
    \caption{
    The comprehensive comparison of the listed methods and \textsc{SeeD}.
    ~\squareSymbol~ represents draft-and-target SD method, while~\triangleSymbol~represents self-draft SD method.
    }
    \label{tab:baseline_comparison}
\end{table*}

\begin{table}[t]
    \centering
    \small
    \resizebox{\columnwidth}{!}{%
    \begin{tabular}{c|c|c|c|c}
    \toprule
    \textbf{LLM Suite}& \textbf{GPUs} & \textbf{Methods} & \textbf{Tokens/s} & \textbf{Speedup} \\
    \midrule
    \multirow{6}{*}{\makecell{\textbf{LLaMA2}\\\textbf{160M/13B}}}& \multirow{3}{*}{\textbf{4$\times$RTX 3090s}} & AR & 38.77 & 1.000$\times$\\
    & & SD & 42.18 & 1.088$\times$\\
    & & \cellcolor{mycell}{\textbf{w. \textsc{SeeD}}} &\cellcolor{mycell}{\textbf{44.93}} & \cellcolor{mycell}{\textbf{1.159$\times$}}\\
    \cmidrule{2-5}
    & \multirow{3}{*}{\textbf{1$\times$RTX A100}} & AR & 39.57 & 1.000$\times$\\
    & & SD & 43.69 & 1.104$\times$\\
    & & \cellcolor{mycell}{\textbf{w. \textsc{SeeD}}} & \cellcolor{mycell}{\textbf{47.25}} & \cellcolor{mycell}{\textbf{1.194$\times$}}\\
    \midrule
    \multirow{6}{*}{\makecell{\textbf{Qwen1.5}\\\textbf{0.5B/7B}}}& \multirow{3}{*}{\textbf{4$\times$RTX 3090s}} & AR & 27.51 & 1.000$\times$\\
    & & SD & 27.43 & 0.997$\times$\\
    & & \cellcolor{mycell}{\textbf{w. \textsc{SeeD}}} & \cellcolor{mycell}{\textbf{29.57}} & \cellcolor{mycell}{\textbf{1.075$\times$}}\\
    \cmidrule{2-5}
    & \multirow{3}{*}{\textbf{1$\times$RTX A100}} & AR & 33.86 & 1.000$\times$\\
    & & SD & 34.91 & 1.031$\times$\\
    & & \cellcolor{mycell}{\textbf{w. \textsc{SeeD}}} & \cellcolor{mycell}{\textbf{39.35}} & \cellcolor{mycell}{\textbf{1.162$\times$}}\\
    \midrule
    \bottomrule
    \end{tabular}
    }
\caption{Speed performance of LLaMA2 suite on Creative Writing dataset under different hardware 
environments with temperture = 1.0 and $k_{\text{config}}$ = (1,1,1,1), as well as the performance of Qwen1.5 suite suite on GSM8K dataset across different hardware environments  with temperture = 1.0 and $k_{\text{config}}$ = (4,2,1).}
\label{table:hardware_performance}
\end{table}

\subsection{Task Performance}
\noindent \paragraph{Accuracy}~\citet{leviathan2023fast} has proved the outputs of AR and SD are the same.
We separately evaluated the performance of the GSM8K dataset using the AR with QWen1.5-7B and \textsc{SeeD}
with the aforementioned QWen1.5 suite using QWen1.5-0.5B and QWen1.5-7B, and found that the performance difference was within $\pm1.5\%$, which is acceptable and substantiates that the performance is effectively \textbf{lossless}.

\noindent \paragraph{Performance on Non-Reasoning Tasks}
\textsc{SeeD} is a versatile method that can be applied not only in reasoning tasks involving TSB but also in non-reasoning tasks.
Its general applicability makes it a robust solution for various scenarios.
We specifically applied \textsc{SeeD} to the TSB in reasoning tasks based on several key considerations:
\begin{itemize}[itemsep=0.5pt, parsep=0pt]
    \item \textbf{Practicality of TSB:} The TSB method allows the generation of multiple sequences simultaneously in both identical and varied input scenarios. This makes it a practical choice for efficient processing.
    \item \textbf{Efficiency on Consumer-Grade GPUs:} Typically, TSB involves generating 2-6 reasoning paths concurrently, which can be handled by consumer-grade GPUs. By contrast, promtping methods like Self-Consistency~\citep{wang2023selfconsistency} often require generating 10-20 sequences, parallelly placing a greater strain on hardware resources.
    \item \textbf{Relevance to Task Difficulty:} Reasoning tasks are challenging benchmarks for evaluating LLMs. If our framework achieves effective acceleration under acceptance in these tasks, it is likely to perform well on simpler tasks, like translation, where the alignment between the target and draft models is better. 
    In early exploratory experiments, \textsc{SeeD} achieved a \textbf{1.31x speedup} over AR on the WMT dataset~\citep{bojar-etal-2014-findings}, demonstrating its efficacy.    
\end{itemize}

\subsection{Hardware Dependency}\label{app:hardware}
The experiments was conducted on a 4$\times$3090 server in the earlier exploratory.
From the experiments on different hardware shown in Figure~\ref{table:hardware_performance}, our method is still effective compared with SD with the same setting.
The speedup performance on 4$\times$3090 is lower than on 1$\times$A100, likely due to the increased communication time between multiple GPUs~\citep{sun2024triforce}.
This is also evident from the Qwen suite results, where SD performs worse than AR on 4$\times$3090.

\subsection{Batch Inference}\label{app:batch_infer}
Batch inference processes multiple sequences of varying lengths.
In SD, each sequence in the same batch requires extra padding due to different acceptance rates and sequence lengths, potentially leading to excessive storage and computation~\citep{ni2024ems}.
This can result in an overly long KV-Cache, thereby slowing down the speedup effect due to inconsistent acceptance lengths.
Our \textsc{SeeD} maintains the original length of KV-Cache without the need for padding based on varying acceptance rates.
Each verified draft sequence corresponds directly to a sequence in the batch (number of draft models $n$ = batch size).
Our parallel drafting approach ensures efficient batch implementation while preserving the acceleration benefits of SD.

\section{Details of Tree Attention}\label{app:tree_attention}
\begin{figure}[t]
    \centering
    \includegraphics[width=0.45\textwidth]{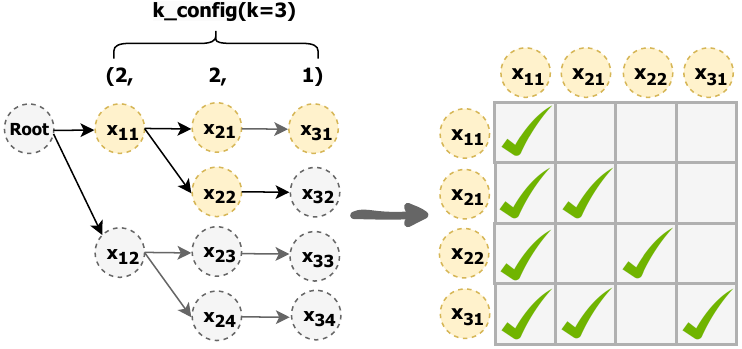}
    \caption{The tree attention used in \textsc{SeeD}, multiple tokens in single sequence concurrently are processed. 
    \textit{Root} indicates previous tokens.
    {\yes} indicates where attention is present, while the rest are masked.
    For simplicity, we only
    visualize the tree attention mask of tokens in \textcolor{yellow}{yellow} colors. 
    }
    \label{fig:tree_attention}
\end{figure}
Setting $k_{\text{config}}$ to (2,2,1) indicates that each draft phase generates a group of $k=3$ tokens, with the first two positions each sampling 2 candidates, and the third position sampling 1.
Figure~\ref{fig:tree_attention} illustrates a case of tree attention with a configuration of $k_{\text{config}}=(2,2,1)$.

\section{Reasoning Tree Construction}
The process of constructing a reasoning tree using the BFS Algorithm is outlined in Algorithm~\ref{alg:seed_bfs}.
\begin{algorithm}[h]
\caption{\textsc{SeeD}($x, p_\theta, \textsc{G}, n, \textsc{E}, s, b$)}
\small
\label{alg:seed_bfs}
\begin{algorithmic}[1]
\State \textbf{Input:} Initial prompt $\mathcal{I}$, speculative scheduled execution with a rounds-scheduled strategy $p_\theta$, thought generator $\textsc{G}(\cdot)$ with a number of thought $n$, states evaluator $\textsc{E}(\cdot)$, step limit $\mathcal{T}$, breadth limit $b$.
\State \textbf{Initialize:} States $S$; $S_0 \gets \{\mathcal{I}\}$
\For{$i = 1, \cdots, \mathcal{T}$}
    \State $S'_i \gets \{[c, z_i] \mid c \gets S_{i-1}$, $z_i \in {\textsc{G}}(p_\theta, c, n)\}$
    \quad \quad \quad \quad \textcolor{DarkCoral}{\Comment{Generate thoughts in Parallel}}
    \State $E_i \gets E(p_\theta, S'_i)$ \textcolor{ForestGreen}{\Comment{Evaluate states in Parallel}}
    \State $S_i \gets \arg \max_{S \subset S'_i, |S| = b} \sum_{s \in S} E_i(s)$
\EndFor \\
\Return $\textsc{G}(p_\theta, \arg \max_{s \in S_\mathcal{T}} E_\mathcal{T}(s), 1)$
\end{algorithmic}
\end{algorithm}

\section{Detailed Setup and Prompts}\label{app:prompts}
We implemented a simple and generic ToT-BFS according to \citet{yao2024tree}.
Within the Thought Generator, we leverage a sampling strategy to generate thoughts for the next thought step. 
Within the State Evaluator, we leverage a value strategy to evaluate the generated thoughts and output a scalar value (\eg, ``1-10'')  or a classification (\eg, ``\textit{good}/\textit{bad}'') which can be heuristically converted into a value.
To introduce diversity in thought generation across all tasks, we set the generation temperature as 0.2/1($\textgreater$0) for the LLaMA suite models and 0.2/0.6/1($\textgreater$0) for the QWen suite models.
The tree depth $\mathcal{T}$ suggests that the operations with varying levels of complexity or iterations, with deeper trees potentially representing more complex calculations or decision-making processes.
The ToT setup of the three tasks \textsc{SeeD} utilized is as follows:
\begin{itemize}[itemsep=0.5pt, parsep=0pt]
    \item \textbf{Creative Writing}: We build a reasoning tree with a depth $\mathcal{T}$ of 2 (with 1 intermediate thought step) that generates 3 plans and passages.
    The State Evaluator assesses the plans and outputs a coherency score with each plan and passage.
    \item \textbf{GSM8K}: We build a reasoning tree with a depth $\mathcal{T}$ of 4 (with 3 intermediate thought steps) that generates 3 sub-questions and corresponding sub-answers.
    This setup aligns with the findings from \citet{hao2023reasoning}, which indicated that three steps are generally sufficient to achieve a passable level of accuracy.
    The State Evaluator assesses them and outputs a number representing the helpfulness for answering the question. We select the one with the highest values and add it to the previous sub-question and sub-answers. 
    \item \textbf{Blocksworld 6-step}: We build a reasoning tree with a depth $\mathcal{T}$ of 7 (with 6 intermediate thought steps) that generates 3 thoughts, including action plans and current actions. 
    Due to the complexity of this task, demonstrations are provided in the prompt, labeled as ``\textit{good}/\textit{bad}'', to assist the State Evaluator in its assessment.
\end{itemize}

The prompts for the tasks described above are presented below.
The \textcolor{orange}{orange} parts in prompts are required for LLM completion.
During the evaluation, we require the LLM to generate both a score and an explanation (a context with 128 new tokens), \textbf{rather than just a score}.
This approach promotes the speedup in generation and makes the evaluation of ToT more reasonable.

\newpage
\begin{tcolorbox}[
enhanced jigsaw,
breakable,
pad at break*=1mm,
colback=white!95!gray,
colframe=gray!50!black,
title={Prompts for GSM8K}]
\small
\textbf{The Thought Generator}
\begin{lstlisting}[breaklines=true, xleftmargin=0pt, breakindent=0pt, columns=fullflexible]
Given a question: (*@\textcolor{blue}{
\{initial\_prompt\}}@*), the previous sub-question and sub-answer is: 
(*@\textcolor{blue}{\{state\_text\}}@*)
Please output the next sub-question to further reason the question.
The sub-question is: (*@\textcolor{orange}{\{sub-question\}}@*)
------------------------------
Given a question: (*@\textcolor{blue}{
\{initial\_prompt\}}@*), the sub-question is: (*@\textcolor{blue}{\{sub\_question\}}@*)
Please answer the sub-question based on the question.
The sub-answer is: (*@\textcolor{orange}{\{sub\_answer\}}@*)
\end{lstlisting}
\textbf{The State Evaluator}
\begin{lstlisting}[breaklines=true, xleftmargin=0pt, breakindent=0pt, columns=fullflexible]
Given a question: (*@\textcolor{blue}{
\{initial\_prompt\}}@*), the sub-question is: (*@\textcolor{blue}{\{sub\_question\}}@*), the sub-answer is: (*@\textcolor{blue}{\{sub\_answer\}}@*)
Please output a number between 1 and 10 to evaluate the answer. The higher the number, the more help there is in answering the question.

The number is: (*@\textcolor{orange}{\{value\}}@*)
\end{lstlisting}
\end{tcolorbox}

\begin{tcolorbox}[
enhanced jigsaw,
breakable,
pad at break*=1mm,
colback=white!95!gray,
colframe=gray!50!black,
title={Prompts for Creative Writing}]
\small
\textbf{The Thought Generator}
\begin{lstlisting}[breaklines=true, xleftmargin=0pt, breakindent=0pt, columns=fullflexible]
Write a coherent passage of 4 short paragraphs. The end sentence of each paragraph must be: (*@\textcolor{blue}{
\{initial\_prompt\}}@*)
Make a plan then write. Your output should be of the following format:

Plan:
Your plan here.

Passage:
Your passage here.

The output is: 
(*@\textcolor{orange}{\{Plan\}}@*) 
(*@\textcolor{orange}{\{Passage\}}@*)

\end{lstlisting}
\textbf{The State Evaluator}
\begin{lstlisting}[breaklines=true, xleftmargin=0pt, breakindent=0pt, columns=fullflexible]
Analyze the passage: (*@\textcolor{blue}{\{Passage\}}@*), then at the last line conclude "Thus the coherency score is [s]", where [s] is an integer from 1 to 10.
The coherency score is: (*@\textcolor{orange}{\{value\}}@*)
\end{lstlisting}
\end{tcolorbox}

\clearpage
\newpage
\begin{tcolorbox}[
breakable,
colback=white!95!gray,
colframe=gray!50!black,
title=Prompts for Blocksworld]
\small
\textbf{The Thought Generator}
\begin{lstlisting}[breaklines=true, xleftmargin=0pt, breakindent=0pt, columns=fullflexible]
I am playing with a set of blocks where I need to arrange the blocks into stacks. Here are the actions I can do:

Pick up a block
Unstack a block from on top of another block
Put down a block
Stack a block on top of another block

I have the following restrictions on my actions:
(*@\textcolor{gray}{\#\#Restrictions on Action\#\#}@*)

(*@\textcolor{gray}{<---Omit demonstrations--->}@*)

[STATEMENT]
(*@\textcolor{blue}{\{initial\_prompt\}}@*)

My plan is as follows:
(*@\textcolor{blue}{\{state\_text\}}@*)
The current action is:
(*@\textcolor{orange}{\{action\}}@*)
\end{lstlisting}
\textbf{The State Evaluator}
\begin{lstlisting}[breaklines=true, xleftmargin=0pt, breakindent=0pt, columns=fullflexible]
I am playing with a set of blocks where I need to arrange the blocks into stacks. Here are the actions I can do:

Pick up a block
Unstack a block from on top of another block
Put down a block
Stack a block on top of another block

I have the following restrictions on my actions:
(*@\textcolor{gray}{\#\#Restrictions on Action\#\#}@*)

(*@\textcolor{gray}{<---Omit demonstrations--->}@*)

Please evaluate whether the given action is a good one under certain conditions.

[STATEMENT]
(*@\textcolor{blue}{\{initial\_prompt\}}@*)
[ACTION]
(*@\textcolor{blue}{\{state\_text\}}@*)
[EVALUATION]
The evaluation is:
(*@\textcolor{orange}{\{evaluation\}}@*)
\end{lstlisting}
\end{tcolorbox}

\newpage
\begin{tcolorbox}[
breakable,
colback=white!95!gray,
colframe=gray!50!black,
title=Restrictions on Action for Blocksworld]
\small{
\begin{lstlisting}[breaklines=true, xleftmargin=0pt, breakindent=0pt, columns=fullflexible]
I have the following restrictions on my actions:
I can only pick up or unstack one block at a time.
I can only pick up or unstack a block if my hand is empty.
I can only pick up a block if the block is on the table and the block is clear. A block is clear if the block has no other blocks on top of it and if the block is not picked up.
I can only unstack a block from on top of another block if the block I am unstacking was really on top of the other block.
I can only unstack a block from on top of another block if the block I am unstacking is clear.
Once I pick up or unstack a block, I am holding the block.
I can only put down a block that I am holding.
I can only stack a block on top of another block if I am holding the block being stacked.
I can only stack a block on top of another block if the block onto which I am stacking the block is clear.
Once I put down or stack a block, my hand becomes empty.
\end{lstlisting}}
\end{tcolorbox}

\end{document}